\documentclass[letterpaper]{article} 
\usepackage{aaai25}  
\usepackage{times}  
\usepackage{helvet}  
\usepackage{courier}  
\usepackage[hyphens]{url}  
\usepackage{graphicx} 
\usepackage{natbib}  
\usepackage{caption} 
\usepackage{algorithm}
\usepackage{algorithmic}
\usepackage{booktabs}
\usepackage{pifont}
\usepackage{multirow}
\usepackage{amsmath}
\usepackage{amssymb}
\usepackage{amsfonts} 
\usepackage{mathrsfs}
\usepackage{newfloat}
\usepackage{listings}
\DeclareCaptionStyle{ruled}{labelfont=normalfont,labelsep=colon,strut=off} 
\floatstyle{ruled}
\newfloat{listing}{tb}{lst}{}
\floatname{listing}{Listing}
\title{Relaxed Rotational Equivariance via $G$-Biases in Vision}
\author {
    Zhiqiang Wu\textsuperscript{\rm 1}\equalcontrib,
    Yingjie Liu\textsuperscript{\rm 1},
    Licheng Sun\textsuperscript{\rm 1}\equalcontrib,
    Jian Yang\textsuperscript{\rm 2},
    Hanlin Dong\textsuperscript{\rm 1},
    Shing-Ho J. Lin\textsuperscript{\rm 3},
    \\
    Xuan Tang\textsuperscript{\rm 4},
    Jinpeng Mi\textsuperscript{\rm 5},
    Bo Jin\textsuperscript{\rm 6},
    Xian Wei\textsuperscript{\rm 1}\footnote{Corresponding author.}
}
\affiliations {
    \textsuperscript{\rm 1}Software Engineering Institute, East China Normal University\\
    \textsuperscript{\rm 2}School of Geospatial Information, Information Engineering University\\
    \textsuperscript{\rm 3}School of Artificial Intelligence, University of Chinese Academy of Sciences\\
    \textsuperscript{\rm 4}School of Communication and Electronic Engineering, East China Normal University\\
    \textsuperscript{\rm 5}Institute of Machine Intelligence, University of Shanghai for Science and Technology\\
    \textsuperscript{\rm 6}	School of Computer Science and Technology, Tongji University\\
    \{51265902095, lcsun\}@stu.ecnu.edu.cn, xian.wei@tum.de 
}

\begin{document}

\maketitle

\begin{abstract}
Group Equivariant Convolution (GConv) can capture rotational equivariance from original data. It assumes uniform and strict rotational equivariance across all features as the transformations under the specific group. However, the presentation or distribution of real-world data rarely conforms to strict rotational equivariance, commonly referred to as Rotational Symmetry-Breaking (RSB) in the system or dataset, making GConv unable to adapt effectively to this phenomenon. Motivated by this, we propose a simple but highly effective method to address this problem, which utilizes a set of learnable biases called $G$-Biases under the group order to break strict group constraints and then achieve a Relaxed Rotational Equivariant Convolution (RREConv). To validate the efficiency of RREConv, we conduct extensive ablation experiments on the discrete rotational group $\mathcal{C}_n$. Experiments demonstrate that the proposed RREConv-based methods achieve excellent performance compared to existing GConv-based methods in both classification and 2D object detection tasks on the natural image datasets.
\end{abstract}

%
\begin{links}
    \link{Code}{https://github.com/wuer5/rrenet}
    \link{Extended version}{https://arxiv.org/abs/2408.12454}
\end{links}

\section{Introduction}
Symmetry prior, such as equivariance, plays a vital role in deep learning \cite{bogatskiy2022symmetrygroupequivariantarchitectures, he2021efficient, esteves2020theoretical, ravanbakhsh2017equivariance}. 
Given the assumption of perfect symmetry in data, recent works on equivariant networks are constrained to operate as strict equivariant or invariant functions.
These works have been shown to learn potential equivariance or symmetry information without additional data, achieving excellent results with improved efficiency and generalization ability \cite{cohen2016groupequivariantconvolutionalnetworks,cohen2016steerablecnns,kondor2018generalizationequivarianceconvolutionneural, Ghosh_2022,kaba2023equivariancelearnedcanonicalizationfunctions,kaba2024symmetrybreakingequivariantneural}.

\begin{figure}[t]
\centering
\includegraphics[width=1\columnwidth]{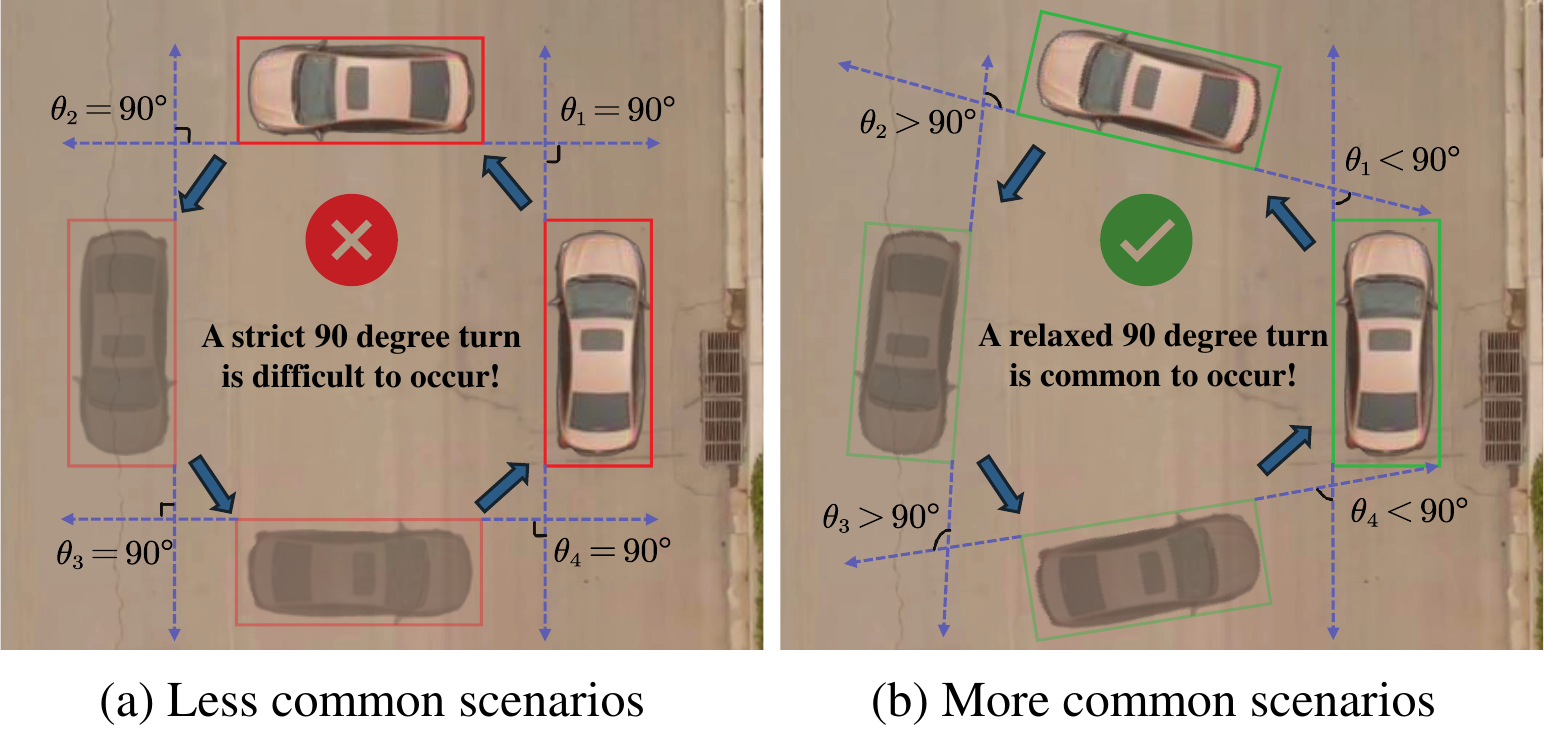}
\caption{(a) A car turning right at an angle of exactly 90 degrees denotes the strict adherence to the motion rules on the group $\mathcal{C}_4$. (b) Another car turning right at an angle of approximately 90 degrees represents a deviation from strict rotational symmetry on the group $\mathcal{C}_4$, leading to Rotational Symmetry-Breaking (RSB) within a car's motion. Note that the figure emphasizes the symmetry of an object's potential motion, not the symmetry of an object itself on the group $\mathcal{C}_4$.
}
\label{fig1}
\end{figure}

However, real-world physical systems seldom conform to perfect or strict symmetry due to the natural laws of absolute motion in the world, a phenomenon commonly termed {Symmetry-Breaking} \cite{wang2022approximately, barone2008symmetry,wang2024discoveringsymmetrybreakingphysical,vernizzi2002rotational, Ghosh_2022, kaba2024symmetrybreakingequivariantneural}. 
In fact, {Symmetry-Breaking} is a broad concept that involves different definitions of objects or systems. Usually, one definition can refer to breaking the symmetry of an object or system itself. Another definition can be the motion of an object or system that cannot precisely conform to the rules of a strict symmetry group (e.g., the typical rotational group $\mathcal{C}_n$), thereby breaking its strict symmetry state.

This paper specifically focuses on the second definition of Symmetry-Breaking in the following context. We observe that this type of Symmetry-Breaking often occurs within an object's potential motion in 2D vision. A common example of breaking the strict symmetry state on the group $\mathcal{C}_4$ within a car can be seen in Figure \ref{fig1}. In (a), a car turns by 90 degrees, which can be challenging to achieve precisely in practical situations. Conversely, in (b), a car turns by more or less than 90 degrees, often with a slightly randomized angle change, reflecting a more realistic depiction of a car’s turning motion in real-world settings. 
Strict rotational symmetry constraints hinder equivariant networks from effectively modelling real-world scenarios and visual perception. 

Following the second definition above, we mainly concentrate on a common case, i.e., Rotational Symmetry-Breaking (RSB) based on GConv \cite{cohen2016groupequivariantconvolutionalnetworks} on the discrete rotational group $\mathcal{C}_n$ in vision domain. 
By rethinking the construction process and principles of GConv, we find that under a strict group $G$, each $G$-transformation convolution filter shares the same copy value, differing only in their positions, which is the key to achieving GConv's strict equivariance. 
For example, GConv cannot effectively capture the equivariance of objects or scenes with rotations other than 90, 180, and 270 degrees on the group $\mathcal{C}_4$.
Therefore, we aim to relax the strict $G$-transformation convolution filter value-sharing problem to adapt to RSB.

\begin{figure}[t]
\centering
\includegraphics[width=1\columnwidth]{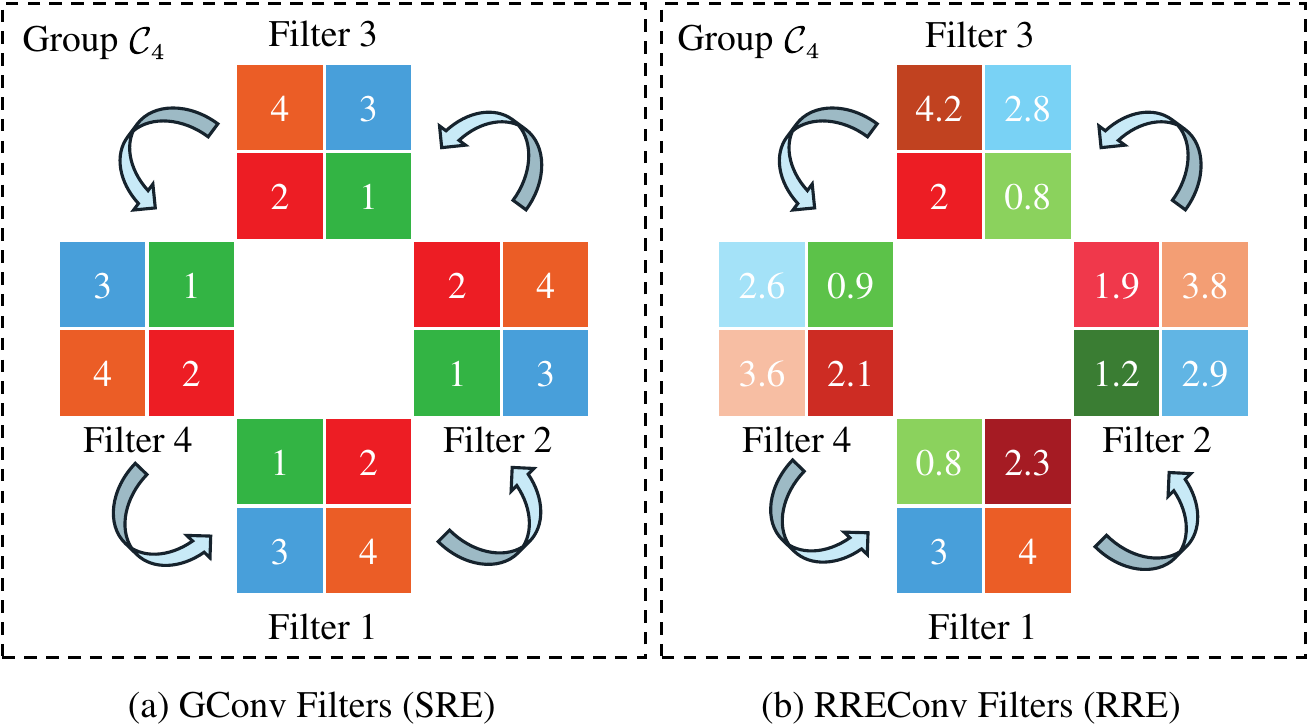}
\caption{The $2 \times 2$ filters between {Strict Rotational Equivariance} (SRE) and {Relaxed Rotational Equivariance} (RRE) on the group $\mathcal{C}_4$. Filters 1 to 4 in Figure (a) have the same values in four directions, whereas Filters 1 to 4 in Figure (b) have slightly different values in four directions.}
\label{fig2}
\end{figure}

Inspired by convolution biases, we innovatively introduce a set of learnable biases called $G$-Biases to add to the $G$-transformation convolution filters. In this paper, we refer to GConv with $G$-Biases as {Relaxed Rotational Equivariant Convolution} (RREConv) on the group $\mathcal{C}_n$. The rotational equivariance of GConv is called {Strict Rotational Equivariance} (SRE), and of RREConv is called {Relaxed Rotational Equivariance} (RRE).
The difference between GConv and RREConv filters can be seen in Figure \ref{fig2}. 
Note that $G$-Biases are learnable parameters that can update end-to-end based on the characteristics or distribution of the dataset.

The main contributions are as follows:
\begin{itemize}
    \item To the best of our knowledge, we are the first to explore RSB within an object's potential motion in vision.
    \item We propose a simple yet efficient method to address RSB based on the existing GConv.
    \item The proposed RREConv enhances the performance of GConv-based models with fewer additional parameters and is easily integrated as a plug-and-play module across various GConv-based models.
\end{itemize}

\section{Related Works}
\subsubsection{Strict Rotational Equivariance (SRE).} SRE is critical for neural networks capturing rotational equivariance from original data, especially in computer vision tasks involving 2D images and 3D objects \cite{marcos2016learning}. As demonstrated by pioneering work \cite{cohen2016groupequivariantconvolutionalnetworks}, introducing group equivariance in traditional CNNs led to the development of novel G-CNNs. Foundational networks such as \cite{li2018deeprotationequivariantnetwork, Rotation-Equivariant-Vector-Field-Networks, chidester2018rotationequivarianceinvarianceconvolutional, weiler2021generale2equivariantsteerablecnns}, have achieved notable success in exploring the rotational equivariance of images. In \cite{veeling2018rotationequivariantcnnsdigital, müller2021rotationequivariantdeeplearningdiffusion}, the application of rotational equivariance in medicine has achieved excellent performance due to the frequent presence of rotational equivariance in medical data. Moreover, rotational equivariance is also prevalent in 2D object detection. For example, ReDet \cite{han2021redetrotationequivariantdetectoraerial} introduces rotational equivariance in detecting aerial objects. MORE-Net \cite{Multi-Oriented-Rotation-Equivariant-Network-for-Object-Detection} proposes the multi-oriented ground objects detector to extract rotation-invariant semantic representations. In addition, EquiSym \cite{seo2022reflectionrotationsymmetrydetection} outputs group equivariant score maps for rotational centres through end-to-end rotational equivariant feature maps. Rotational equivariance has also seen significant advancements in 3D vision. In \cite{weiler20183dsteerablecnnslearning, fuchs2020se3transformers3drototranslationequivariant}, 3D Steerable CNNs and SE(3)-Transformers are proposed to explore 3D rotational equivariance. In 3D object detection, 3D equivariance remains crucial. DuEqNet \cite{wang2023dueqnetdualequivariancenetworkoutdoor} focuses on introducing 3D rotation group equivariance constraints in 3D object detection tasks, highlighting the importance of rotational equivariant features in enhancing the robustness of 3D detection models. ProEqBEV \cite{ProEqBEV} demonstrates the effectiveness of rotational equivariant BEV features. These methods preserve the structural integrity of the learned features under rotation, enhancing the robustness of the network to such strict rotational transformations.

\subsubsection{Relaxed Rotational Equivariance (RRE).} 
However, the methods of SRE mentioned above make them poorly suited for handling RSB in both 2D and 3D fields, as they assume that the rotational equivariance exhibits perfect rotational symmetry, which is a rare case in real-world scenarios \cite{wu2015flip, dieleman2016exploiting, kavukcuoglu2009learning}. In contrast, explorations into RRE reveal a significant gap in current research. Although SRE models are effective under ideal conditions, they fall short when dealing with the more common, imperfect symmetries in the real world. In some theoretical works \cite{kaba2024symmetrybreakingequivariantneural, kaba2023equivariancelearnedcanonicalizationfunctions}, they meticulously analyze the importance and potential application scenarios of relaxed equivariance. As highlighted in \cite{romero2023learningpartialequivariancesdata}, partial rotational equivariance is more effective in representing real-world data than full rotational equivariance. Similarly, \cite{vanderouderaa2022relaxingequivarianceconstraintsnonstationary} emphasizes that the constraints in equivariance can be too restrictive. In addition, \cite{wang2024discoveringsymmetrybreakingphysical} explores the relaxed equivariance of physical systems. These methods confirm that relaxed equivariance (e.g., especially RRE) is more suitable for real-world scenarios.

\subsubsection{Rotational Symmetry and Symmetry-Breaking.} 
Rotational symmetry \cite{li2024enforcing}, a broad concept in natural systems, refers to objects or patterns that retain their appearance when rotated at specific angles \cite{barone2008symmetry}. Although prevalent in theoretical models, this ideal or pristine form of symmetry is often disrupted in real-world scenarios, a phenomenon referred to as RSB \cite{vernizzi2002rotational}. Recent progress in understanding these deviations has spurred the creation of new methodologies aimed at tackling these challenges. For instance, methods suggested by \cite{desai2022symmetry} propose techniques to integrate Symmetry-Breaking elements into models, enhancing their performance in data mining and analysis tasks. However, most researchers focus on strict rotational symmetry and often overlook the phenomenon of RSB. Therefore, this paper primarily focuses on addressing this phenomenon.

\section{Preliminary}
\subsubsection{Definition of Strict Equivariance.} Assume that the input group representation $\varphi_{X}$ of $G$ acts on $X$ and the output group representation $\varphi_{Y}$ of $G$ acts on $Y$. A learnable function $f_{\mathrm{strict}}: X \to Y$ satisfies {Strict Equivariance} if
\begin{equation}
f_{\mathrm{strict}}(\varphi_{X}(g)(\mathbf{x}))=\varphi_{Y}(g)f_{\mathrm{strict}}(\mathbf{x}),
\end{equation}
where all $\mathbf{x} \in X$ and $g \in G$. 
\subsubsection{Definition of $\varepsilon$-Relaxed or $\varepsilon$-Approximate Equivariance.} Consistent with the above definition, a learnable function $f_{\mathrm{relaxed}}: X \to Y$ satisfies {Relaxed Equivariance} if
\begin{equation}
\| f_{\mathrm{relaxed}}(\varphi_{X}(g)(\mathbf{x})) - \varphi_{Y}(g)f_{\mathrm{relaxed}}(\mathbf{x}) \| \le \varepsilon ,  
\end{equation}
where all $\mathbf{x} \in X$ and $g \in G$. The upper bound $\varepsilon$ is usually a small number, with a relatively larger value indicating the greater level of relaxation and a relatively smaller value showing the stronger level of equivariance. Especially when $\varepsilon = 0$, we have $f_{\mathrm{relaxed}} = f_{\mathrm{strict}}$. Note that $\varepsilon$ is determined by the level of Symmetry-Breaking of the dataset or system, which is an implicit constant.
\subsubsection{Strict Equivariant Network.} 
Given a set of strict equivariant functions $\{f_i\}$, a strict equivariant network can be the composition function of $\{f_i\}$. Assume $f_1$ and $f_2$ satisfy strict equivariance and $g_1, g_2 \in G$, then their composition $f_1 \circ f_2 = f_1(f_2(\cdots))$ also satisfies strict equivariance. Since $f_1(g_1\cdot x)=g_1 \cdot f_1(x)$ and $f_2(g_2\cdot x)=g_2 \cdot f_2(x)$, we have $f_2(f_1(g_1 \cdot x))=f_2(g_1 \cdot f_1(x))=g_1 \cdot f_2(f_1(x))$, completing the proof.
The challenge of a strict equivariant network is in designing equivariant layers. Two typical methods are raised by weight sharing \cite{cohen2016groupequivariantconvolutionalnetworks} and weight tying \cite{cohen2016steerablecnns}.

\subsubsection{Relaxed Equivariant Network.} A strict equivariant network considers equivariance but assumes uniform and strict equivariance from the original data. However, real-world data rarely conforms to strict equivariance. To address this problem, we relax the $G$-transformation filters to realize relaxed equivariance. Also, like above, given a set of relaxed equivariant functions $\{\tilde{f_i}\}$, a relaxed equivariant network can be the composition function of $\{\tilde{f_i}\}$. The proof can be referred to \cite{kaba2024symmetrybreakingequivariantneural}.

\subsection{Group Equivariant Convolution (GConv)} 
GConv achieves equivariant inductive biases by sharing weights convolution filters under group transformations. In a special case, CNNs achieve translation equivariance through translation transformations on the plane $\mathbb{Z}^2$.

To begin, we define the group operator $\varphi_G(\mathbf{\cdot})$ performs the $G$-transformation in the last two dimensions and cyclical permutations in the input channel dimension for $\cdot$, and the symbol $[ \ \cdots \ ]$ denotes the Pytorch style index operation. For convenience, we also define $C_{l}$, $k_l$, $h_l$, and $w_l$ denote the channel number, filter size, width, and height of the 2D input or output in the $l$-layer, respectively. These definitions are used in the following context.

\subsubsection{Lift Convolution.}
The first layer of G-CNNs typically lifts the input on the plane $\mathbb{Z}^2$ to the group $G$. Assume the input $\mathcal{Y}_1$ of size $[C_1, h_1, w_1]$ and the initial weight $\mathcal{W}_1$ with {Kaiming Distribution} of size $[C_2, C_1, k_1, k_1]$ on the plane $\mathbb{Z}^2$ in the first layer.
Therefore, we obtain the full lift convolution filter $\mathcal{F}_1 = \varphi_G(\mathcal{W}_1)$ of size $[C_2, G_2, C_1, k_1, k_1]$ that contains an additional dimension $G_2$ for the output group. 
Note that $\mathcal{F}_1$ is constructed from $\mathcal{W}_1$ during each forward function.
For all $ u \in [1, C_2], v \in [1, G_2]$, the lift convolution can be performed by convolving over the input channel $C_1$ and summing up the outputs as follows: 
\begin{equation}
\mathcal{Y}_2[u,v,:,:]  = \sum_{m}^{C_1} \mathcal{Y}_1[m,:,:] \ast \mathcal{F}_1[u,v,m,:,:], 
\end{equation}
with the size of the output $\mathcal{Y}_2$ is $[C_2, G_2, h_2, w_2]$.
\subsubsection{Group Convolution.}
Unlike the input on the plane $\mathbb{Z}^2$, GConv typically encodes the added group $G$ in an extra tensor dimension. 
Assume the input $\mathcal{Y}_l$ of size $[C_l, G_l, h_l, w_l]$ on the group $G$, where $G_l$ denotes the dimension of $G$ in the $l$-layer ($l \geq 2$), and the initial weight $\mathcal{W}_l$ with {Kaiming Distribution} of size $[C_{l+1}, C_l, G_l, k_l, k_l]$ contains an additional dimension $G_l$ for the input group. 
Then, we obtain the full group convolution filter $\mathcal{F}_l = \varphi_G({\mathcal{W}_l})$ of size $[C_{l+1}, G_{l+1}, C_l, G_l, k_l, k_l]$ containing an additional dimension $G_{l+1}$ for the output group. 
For all $ u \in [1, C_{l+1}], v \in [1, G_{l+1}]$, the group convolution can be performed by convolving over the input channel $C_l$ and input group dimension $G_l$, and summing up the outputs as follows: 
\begin{equation}
\mathcal{Y}_{l+1}[u,v,:,:]=\sum_{m}^{C_l} \sum_{n}^{G_l} \mathcal{Y}_l[m,n:,:] \ast \mathcal{F}_l [u,v,m,n, :,:], 
\end{equation}
with the size of the output $\mathcal{Y}_{l+1}$ is $[C_{l+1}, G_{l+1}, h_{l+1}, w_{l+1}]$.
Since the group convolution is the function $f: G \to G$, we have $G_{l+1}=G_{l}=\text{Dim}(G)$, where the $\text{Dim}(G)$ denotes the dimension of $G$ (e.g., $2$ / $4$ / $8$ on the group $\mathcal{C}_2$ / $\mathcal{C}_4$ / $\mathcal{C}_8$).

\begin{figure*}[htpb]
\centering
\includegraphics[width=2\columnwidth]{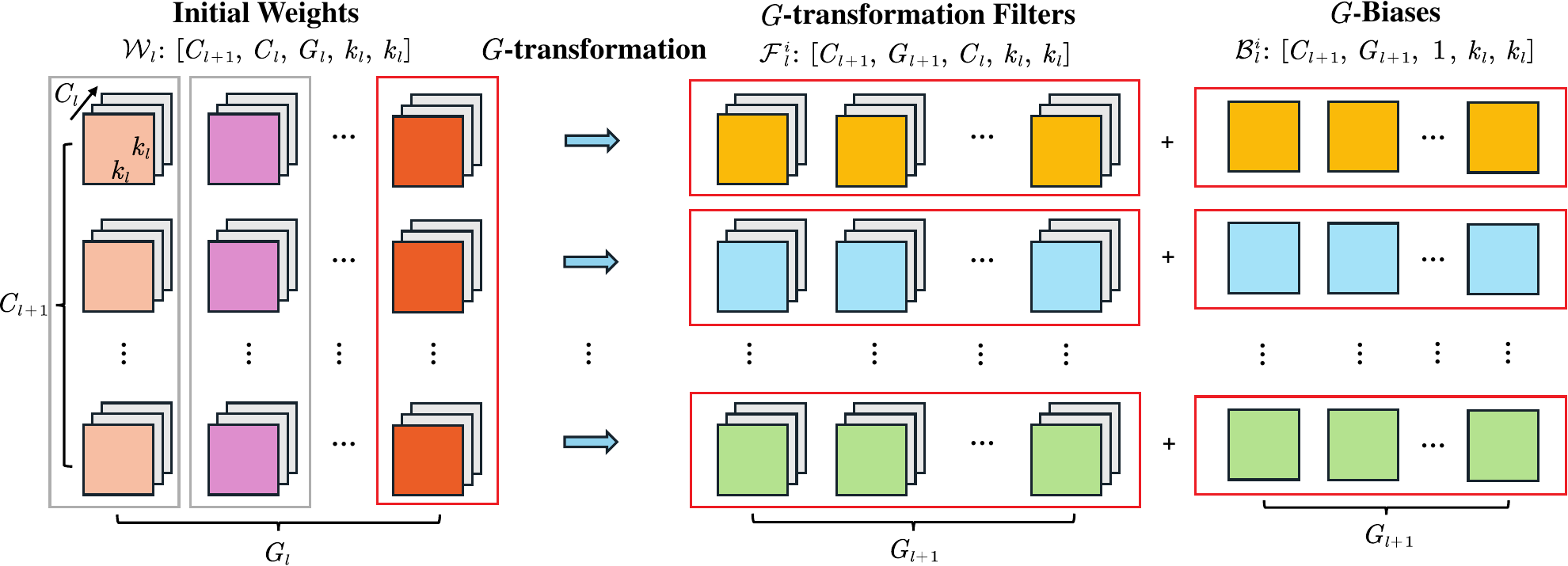}
\caption{The construction of {Relaxed Rotational Equivariant Filter} (RREF). Note that the initial weights have $G_l$ columns, but only the $G$-transformation in the last column (\textbf{Red Box}) is shown here for the convenience of drawing. The operations in other columns (\textbf{Gray Box}) are the same as the last column (\textbf{Red Box}).}
\label{fig3}
\end{figure*}

\section{Proposed Method}

We focus on a specific case, i.e., Relaxed Rotational Equivariance (RRE) for Rotational Symmetry-Breaking (RSB) on the discrete rotational group $\mathcal{C}_n$. 
Then, we introduce the method of the proposed Relaxed Rotational Equivariant Conolution (RREConv) and the model for both classification and detection tasks based on RREConv in 2D vision.

\subsection{Relaxed Rotational Equivariant Filter (RREF)} 
The construction of RREF is the key to our method. 
Assume an initial weight $\mathcal{W}_l$ of size $[C_{l+1}, C_{l}, G_l, k_l, k_l]$ with {Kaiming Distribution} on the group $G$ in the $l$-layer, where $G = \mathcal{C}_n$ and $G_l = n$ especially. 
Define a set of affine matrices $\mathcal{A}=\{\mathcal{A}_i \mid i \in \{0, 1, \cdots, n-1\}\}$ for the $G$-transformation, where  
\begin{equation}
    \mathcal{A}_i = 
    \begin{bmatrix}
    \cos \left({2\pi i}/{n} \right)& -\sin\left ({2\pi i}/{n} \right )\\
    \sin\left ({2\pi i}/{n} \right )&\cos\left ({2\pi i}/{n} \right )\\
\end{bmatrix}.
\end{equation}
Before $G$-transformation for coordinates, assume the coordinate system is at the centre of the 2D plane and define a function $\texttt{CoorSet}(\cdot)$ to obtain the set of all coordinates of $\cdot$.
For all $u \in [1, C_{l+1}], m \in [1, C_l], n \in [1, G_l]$, 2D coordinate pair $(\mathbf{x},\mathbf{y} ) \in \texttt{CoorSet}({\mathcal{W}_l}[u,m,n,:,:])$, we can obtain new 2D coordinate pair $(\tilde{\mathbf{x}}, \tilde{\mathbf{y}})$ after $G$-transformation for coordinates on $\mathcal{W}_l$ as follows:
\begin{equation}
    (\tilde{\mathbf{x}_i}, \tilde{\mathbf{y}_i})^\top = \mathcal{A}_{i}\left(\mathbf{x}, \mathbf{y}\right)^\top, \ \ \forall i \in \{0, 1, \cdots, n-1\}
\end{equation}
Now, we obtain the $i$-filter of SRE on $G$ as follows:
\begin{equation}
    \mathcal{F}_l^i[u,m,n,\tilde{\mathbf{x}_i}, \tilde{\mathbf{y}_i}]=\mathcal{W}_l[u,m,n,\mathbf{x},\mathbf{y}],
\end{equation}
if $(\tilde{\mathbf{x}_i}, \tilde{\mathbf{y}_i}) \in \texttt{CoorSet}({\mathcal{F}_l}[u,m,n,:,:])$. 
Note that some out-of-bounds coordinates of $\mathcal{F}_l^i$ may occur in some groups (e.g., $\mathcal{C}_6$ and $\mathcal{C}_8$) except $\mathcal{C}_2$ and $\mathcal{C}_4$. Based on this reason, some coordinates remain unassigned after the $G$-transformation from $\mathcal{W}_l$. For these coordinates, we employ {Bilinear Interpolation}.
To achieve RREF, we introduce a set of learnable $G$-Biases $\mathcal{B}_l = \{\mathcal{B}^i_l \mid i\in \{0, 1, \cdots, n-1\}\}$, where the size of $\mathcal{B}^i_l$ is $[C_{l+1}, 1, 1, k_l,k_l]$ with {Zero Distribution} in the $l$-layer. 
Note that $\mathcal{B}_l$ can be updated end-to-end during the training period to adapt RSB in the dataset. Thus, the final values of $\mathcal{B}_l$ in $l$-layer are determined by datasets.
Then, we obtain the $i$-filter of RRE on $G$ as follows:
\begin{equation}
      \mathcal{R}_l^i[u,m,n,:,:]=\mathcal{F}^i_l[u,m,n,:,:] + \mathcal{B}_l^i[u,1,1,:,:].
\end{equation}
Thus, the full RREF in the $l$-layer can be stacked as follows:
\begin{equation}
    \mathcal{R}_l = Stack(\{\mathcal{R}_l^i \mid i \in \{0, 1, \cdots n-1\}\}),
\end{equation}
with the size of $[C_{l+1}, G_{l+1}, C_l, G_l, k_l,k_l]$. An easily understandable construction of RREF can be seen in Figure \ref{fig3}.
\subsection{Relaxed Rotational Equivariant Convolution}
Consistent with the operator in Eq. (3) and Eq. (4), our {Relaxed Rotational Lift Convolution} (RRLConv) and {Relaxed Rotational Equivariant Convolution} (RREConv) can be written as Eq. (10) and Eq. (11) below, respectively:
\begin{equation}
\mathcal{Y}_2[u,v,:,:]  = \sum_{m}^{C_1} \mathcal{Y}_1[m,:,:] \ast \mathcal{R}_1 [u,v,m,:,:].
\end{equation}
\begin{equation}
\mathcal{Y}_{l+1}[u,v,:,:]=\sum_{m}^{C_l} \sum_{n}^{G_l} \mathcal{Y}_l[m,n:,:] \ast \mathcal{R}_l [u,v,m,n,:,:].
\end{equation}
\subsection{Proof and Analysis}
\subsubsection{Conclusion 1.} We say RREConv satisfies Eq. (2). 
\subsubsection{Proof 1.} On the group $G=\mathcal{C}_n$, we define the operator $\varphi_G^j(\cdot)$ that rotates $\cdot$ by $2\pi i/n$. For any $j \in \{0, 1, \cdots, n-1\}$, we have two calculations as follows:
\begin{equation}
\begin{split}
\mathcal{Y}_{l+1}(\varphi_G^j(\mathcal{Y}_l))=\sum_{m}^{C_l} \sum_{n}^{G_l} \varphi_G^j(\mathcal{Y}_l) \ast \mathcal{R}_l
\\
=\sum_{m}^{C_l} \sum_{n}^{G_l} \varphi_G^j(\mathcal{Y}_l) \ast \mathcal{F}_l  + \sum_{m}^{C_l} \sum_{n}^{G_l}\varphi_G^j(\mathcal{Y}_l) \ast \mathcal{B}_l.
\end{split}
\end{equation}
\begin{equation}
\begin{split}
\varphi_G^j(\mathcal{Y}_{l+1}(\mathcal{Y}_l))=\sum_{m}^{C_l} \sum_{n}^{G_l} \varphi_G^j(\mathcal{Y}_l) \ast \varphi_G^j(\mathcal{R}_l)
\\
=
\sum_{m}^{C_l} \sum_{n}^{G_l} \varphi_G^j(\mathcal{Y}_l) \ast \varphi_G^j(\mathcal{F}_l) + \sum_{m}^{C_l} \sum_{n}^{G_l} \varphi_G^j(\mathcal{Y}_l) \ast \varphi_G^j(\mathcal{B}_l).
\end{split}
\end{equation}
Since $\mathcal{C}_n$ is a cyclic group, $\forall i \in \{0,1,\cdots,n-1\} $, we have:
\begin{equation}
\begin{split}
\varphi_G^j(\{\mathcal{F}_l^i\}) = \{\varphi_G^j(\mathcal{F}_l^i)\} = \{\mathcal{F}_l^{(i+j)~\%~n}\}
= \{\mathcal{F}_l^i\},
\\
\text{then, }
 \sum_{m}^{C_l} \sum_{n}^{G_l} \varphi_G^j(\mathcal{Y}_l) \ast \varphi_G^j(\mathcal{F}_l)=
 \sum_{m}^{C_l} \sum_{n}^{G_l} \varphi_G^j(\mathcal{Y}_l) \ast \mathcal{F}_l.
\end{split}
\end{equation}
Then we have the L2-Norm as follows:
\begin{equation}
\begin{split}
\|\mathcal{Y}_{l+1}(\varphi_G^j(\mathcal{Y}_l)) - \varphi_G^j(\mathcal{Y}_{l+1}(\mathcal{Y}_l))\|=
\\
\sum_{m}^{C_l} \sum_{n}^{G_l} \varphi_G^j(\mathcal{Y}_l) \ast (\mathcal{B}_l - \varphi_G^j(\mathcal{B}_l))\leq \varepsilon.
\end{split}
\end{equation}
Thus, the proposed RREConv satisfies Eq. (2). Especially when $\mathcal{B}_l=0$, Eq. (15) equals $0$ (i.e., $\varepsilon=0$), where RREConv satisfies Eq. (1). Note that $\mathcal{B}_l$ updates end-to-end. Only at the beginning of the training period, $\mathcal{B}_l=0$.
\subsubsection{Projection Error.}
Considering Eq. (11) of RREConv and Eq. (4) of GConv, we can obtain the {Projection Error} (PE):
\begin{equation}
\begin{split}
\|
\sum_{m}^{C_l} \sum_{n}^{G_l} (\mathcal{Y}_l \ast \mathcal{R}_l - \mathcal{Y}_l \ast \mathcal{F}_l)
\|
=
\|
\sum_{m}^{C_l} \sum_{n}^{G_l} \mathcal{Y}_l \ast \mathcal{B}_l
\|.
\end{split}
\end{equation}
Therefore, each RREConv in the RRE network aims to optimize PE to adopt RSB in natural image datasets.

\begin{figure}[htpb]
\centering
\includegraphics[width=1\columnwidth]{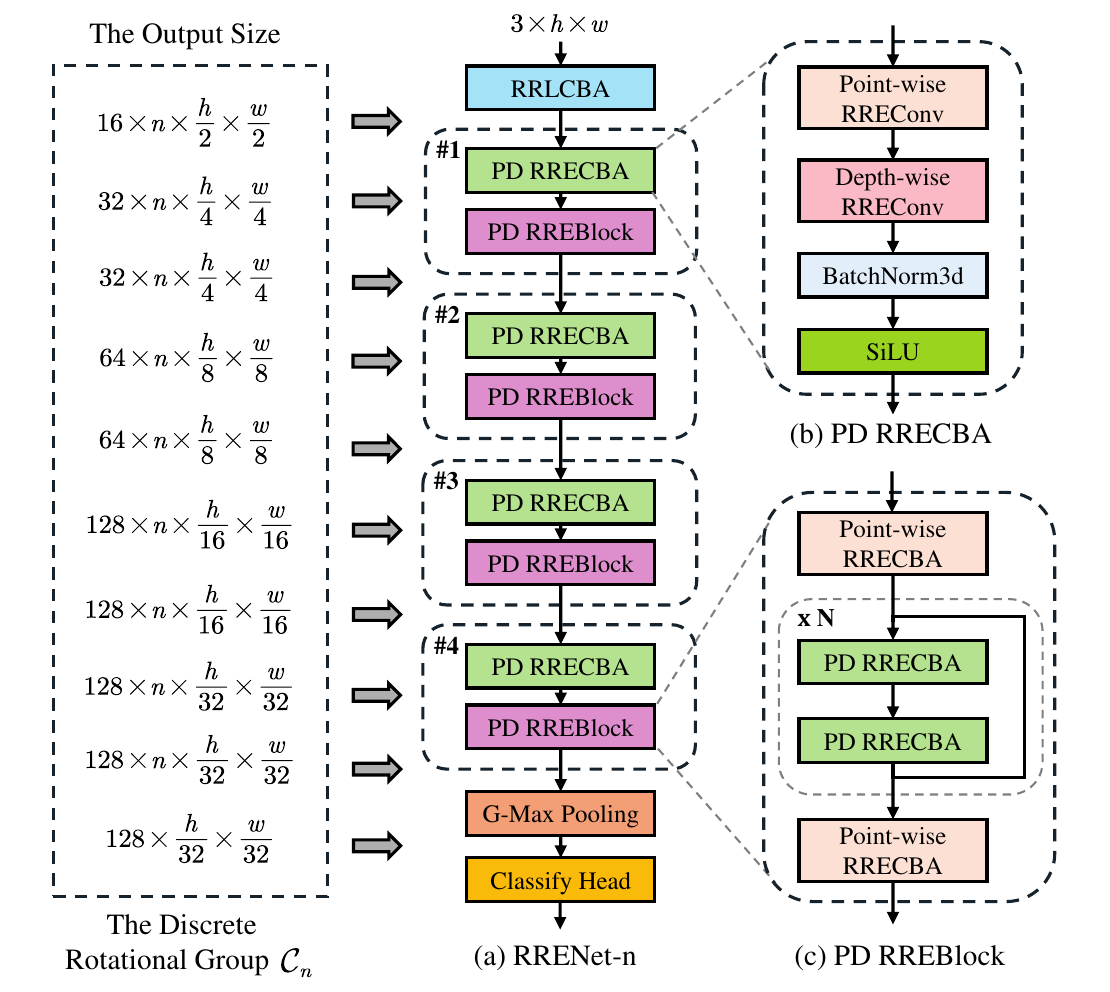}
\caption{The architecture of the backbone RRENet-n based on RREConv. Note that $n$ denotes the dimension of $\mathcal{C}_n$, and "-CBA" means "Conv + BatchNorm + Activate" operations.}
\label{fig4}
\end{figure}

\subsection{Model Architecture} 
\subsubsection{Relaxed Rotational Equivariance Network (RRENet).}
We propose the RRENet based on RREConv. Considering significant computational and parameter overhead caused by the additional $G$-dimension, we redesign the Point-wise and Depth-wise versions of RREConv. We adopt the classic structure of ResNet \cite{he2016deep}, where each PD RREBlock consists of two Point-wise RRECBAs adjusting the number of channels, with two PD RRECBAs in between, and residual connections are used, as shown in Figure \ref{fig4}. 

\subsubsection{Relaxed Rotational Equivariance Detector (RREDet).}
\begin{figure}[htpb]
\centering
\includegraphics[width=1\columnwidth]{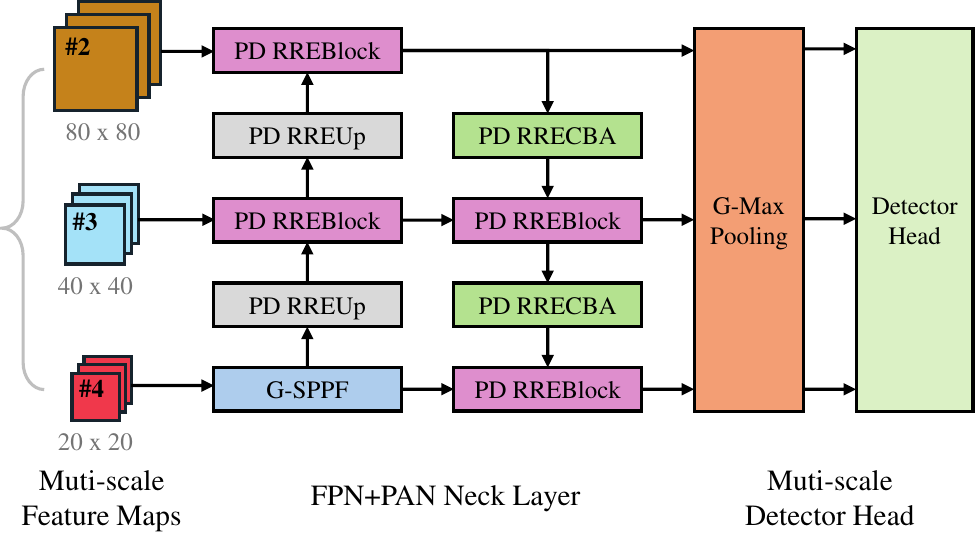}
\caption{The architecture of RREDet. Note that \#\{2,3,4\} denote muti-scale feature maps from \{2,3,4\}-layer in the backbone RRENet. The PD RREUp adopts the same structure as the PD RREConv, except for transposed convolution.}
\label{fig5}
\end{figure}

We also propose the RREDet, where we use the RRENet as the backbone to obtain the initial three scale features in \{2, 3, 4\}-layer, and the FPN+PAN architecture as the neck layer to obtain the final three scale features with size $80 \times 80$, $40 \times 40$, and $20 \times 20$ for different-size object detection. 
Note that the feature maps in 4-layer are input to the G-SPPF for Spatial Pyramid Pooling~\cite{SPP} on the group $\mathcal{C}_n$.
Consequently, these features are fed into the G-Max Pooling and the Detector Head, as shown in Figure \ref{fig5}.

\section{Experiments}
In this section, we conduct extensive ablation experiments to demonstrate the effectiveness of the proposed RREConv in both classification and 2D object detection tasks. All the parameters are set the same, and all the experiments are conducted on the dual RTX-4090 GPUs. 

We evaluate the proposed method on the CIFAR10 / 100, PASCAL VOC07+12, and MS COCO2017 datasets. 
The ablation experiments mentioned above on both tasks show that the proposed method for RRE with fewer additional parameters achieves reasonable performance growth compared to the method for SRE. Also, the proposed method does not increase training and inference time.

\subsection{Training Details}
All training in this paper is based on the famous engine library "Ultralytics"~\cite{Jocher_Ultralytics_YOLO_2023}.

\subsubsection{Data Augmentation.} Like Ultralytics, we employ the default data augmentation settings such as Erasing of 0.4, Mosaic of 1.0, HSV-Saturation of 0.7, HSV-Value of 0.4, and HSV-Hue of 0.015 for the 2D object detection tasks. We also use random augmentation for classification tasks.

\begin{table*}[t]
\centering
\begin{tabular}{cclclcllc}
\toprule
\multirow{2}*{Group $G$} & \multirow{2}*{Type of Equivariance}  & \multicolumn{2}{c}{CIFAR-10} & \multicolumn{2}{c}{CIFAR-100} & \multicolumn{3}{c}{PASCAL VOC07+12} \\ \cmidrule(lr){3-4} \cmidrule(lr){5-6} \cmidrule(lr){7-9}
& & Top-1 Acc. & \#Param. & Top-1 Acc. & \#Param. & $\text{AP}_{\text{50}}$ & $\text{AP}_{\text{50:95}}$ & \#Param. 
\\
\midrule

$\mathbb{Z}^2$ & NRE & $95.6$ & $1.08$M & $77.2$ & $1.19$M & $79.1$ & $58.6$ & $3.11$M
\\
\midrule
\multirow{2}*{$\mathcal{C}_2$} & SRE & $94.4_{{ base}}$ & $0.49$M & $76.1_{{ base}}$ & $0.61$M & $78.9_{{ base}}$ & $58.3_{{ base}}$ & $1.81$M \\
& RRE & $\textbf{95.0}_{{ +0.6}}$ & $0.51$M & $\textbf{76.6}_{{ +0.5}}$ & $0.63$M & $\textbf{80.1}_{{ +1.2}}$ & $\textbf{59.7}_{{ +1.4}}$ & $1.85$M
\\
\midrule
\multirow{2}*{$\mathcal{C}_4$} & SRE &$95.6_{{ base}}$ & $0.79$M & $80.1_{{ base}}$ & $0.91$M & $83.1_{{ base}}$ & $64.1_{{ base}}$ & $2.83$M \\ 
& RRE & $\textbf{96.5}_{{ +0.9}}$ & $0.83$M & $\textbf{80.9}_{{ +0.8}}$ & $0.95$M & $\textbf{84.1}_{{ +1.0}}$ & $\textbf{65.2}_{{ +1.1}}$ & $2.91$M
\\
\midrule
\multirow{2}*{$\mathcal{C}_6$} & SRE & $96.0_{{ base}}$ & $1.09$M & $80.6_{{ base}}$ & $1.21$M & $83.8_{{ base}}$ & $64.7_{{ base}}$ & $3.85$M\\
& RRE & $\textbf{96.8}_{{ +0.8}}$ & $1.16$M & $\textbf{81.3}_{{ +0.7}}$ & $1.27$M & $\textbf{84.5}_{{ +0.7}}$ & $\textbf{65.5}_{{ +0.8}}$ & $3.98$M
\\
\midrule
\multirow{2}*{$\mathcal{C}_8$} & SRE & $96.5_{{ base}}$ & $1.39$M & $82.1_{{ base}}$ & $1.39$M & $85.2_{{ base}}$ & $66.6_{{ base}}$ & $4.88$M\\
& RRE & $\textbf{97.2}_{{ +0.7}}$ & $1.48$M & $\textbf{82.7}_{{ +0.6}}$ & $1.48$M & $\textbf{86.0}_{{ +0.8}}$ & $\textbf{67.5}_{{ +0.9}}$ & $5.04$M
\\
\bottomrule
\end{tabular}
\caption{Ablation experiments compared with NRE, SRE, and RRE on the group $\mathcal{C}_n (n=2,4,6,8)$  on the CIFAR10 / 100 datasets and the PASCAL VOC07+12 dataset. All models are trained from scratch on the architecture of the proposed RRENet for classification tasks and the proposed RREDet for 2D object detection tasks.}
\label{tab1}
\end{table*}

\subsubsection{Training Settings.}
All models are trained from scratch for 300 epochs for the 2D object detection tasks and 100 epochs for the classification tasks. We also default to using an SGD optimizer for both tasks with an initial learning rate of 0.01, a final learning rate of 0.01, a momentum of 0.937, a weight decay of 5e-4, a warmup epoch of 3, a warmup momentum of 0.8, and a warmup bias learning rate of 0.1.

\subsection{Experimental Results}
\subsubsection{Ablation Experiments in Classification.} To evaluate the effectiveness of the proposed RREConv, we conduct extensive ablation experiments on the CIFAR10 / 100 datasets, using the architecture of RRENet with {Non-Rotational Equivariance} (NRE) on the plane $\mathbb{Z}^2$, with SRE and RRE on the group $\mathcal{C}_n$. For SRE, we replace all RREConv with GConv. For NRE, we replace all RREConvs with vanilla Convs but remove G-Max Pooling. We keep all training parameters consistent in NRE, SRE, and RRE. 

The results can be seen in Table \ref{tab1}. 
From the table, the top-1 accuracy of the model with RRE consistently outperforms their NRE and SRE counterparts on the $\mathcal{C}_4$, $\mathcal{C}_6$, and $\mathcal{C}_8$ groups. On the group $\mathcal{C}_2$, although the top-1 accuracy of the models with RRE and SRE is lower than that with NRE, their parameters are only half of the model with NRE. The results demonstrate that RRE achieves better results in classification tasks while maintaining a minor parameter increase compared to SRE. In addition, we find that the model with SRE or RRE on the group $\mathcal{C}_4$ achieves the trade-off between parameters and performance.

\subsubsection{Ablation Experiments in 2D Object Detection.} We also conduct extensive ablation experiments on the PASCAL VOC07+12 dataset to validate the effectiveness of the proposed RREConv in 2D object detection tasks using the standard Average Precision (AP) metric. We typically employ the $\text{AP}_{\text{50}}$ metric at an Intersection over Union (IoU) threshold of $0.5$, and the $\text{AP}_{\text{50:95}}$ metric across IoU thresholds ranging from $0.5$ to $0.95$ as key evaluation metrics.  

As shown in Table \ref{tab1}, the $\text{AP}_{\text{50}}$ and $\text{AP}_{\text{50:95}}$ of the model with RRE consistently surpasses their NRE and SRE counterparts on all groups. The results prove that the proposed method for RRE can also achieve better results in 2D object detection tasks and maintain less parameter increase compared to the method for RRE. Like the results in classification tasks, the models with SRE or RRE on the group $\mathcal{C}_4$ also balance parameters and performance.

\subsubsection{Comparison with other models in Classification.} 
Table \ref{tab2} presents comparative results of three different-size RRENet models (i.e., RRENet-n / m / s) on the group $\mathcal{C}_4$ against other classic convolutional models, including WideResNet, ResNeXt-29, DenseNet-BC, and Res2NeXt-29, on the CIFAR-100 dataset. 
From the table, RRENet-m ($\mathcal{C}_4$) achieves a significant improvement in the top-1 accuracy, with a range of $2.2\%$-$6.9\%$ enhancement over others, but its parameters are only $13\%$-$34\%$ of others. RRENet-s ($\mathcal{C}_4$) still surpasses others, but it has fewer parameters, only one-third that of RRENet-m ($\mathcal{C}_4$), achieving a better balance between parameters and accuracy. Although RRENet-n ($\mathcal{C}_4$) only exceeds WideResNet over others, its parameters are only $0.95$M, which is only $2.6\%$ that of WideResNet.

\begin{table}[t]
\centering
\begin{tabular}{lcc}
\toprule
Model & Top-1 Acc. & \#Param.
\\
\midrule
WideResNet & $79.5$ & $36.5$M
\\
ResNeXt-29 & $82.7$ & $68.1$M
\\
DenseNet-BC & $82.8$ & $25.6$M
\\
Res2NeXt-29 & $83.2$ & $36.7$M
\\
\midrule
RRENet-n ($\mathcal{C}_4$) & $80.9$ & $0.95$M
\\
RRENet-s ($\mathcal{C}_4$) & $83.5$ & $2.97$M
\\
RRENet-m ($\mathcal{C}_4$) & $\textbf{85.0}$ & $8.66$M
\\
\bottomrule
\end{tabular}
\caption{Top-1 Accuracy ($\%$) on the CIFAR-100 dataset. All models are trained from scratch.}
\label{tab2}
\end{table}

\begin{table}[htpb]
\centering
\begin{tabular}{lccr}
\toprule
Model & $\text{AP}_{\text{50}}$ & $\text{AP}_{\text{50:95}}$ & \#Param.
\\
\midrule
YOLOv8-n & $78.6$ & $57.5$ & $3.0$M
\\
YOLOv8-s & $81.6$ & $61.6$ & $11.1$M
\\
YOLOv8-m & $83.7$ & $65.3$ & $25.9$M
\\
\midrule
RREDet-n ($\mathcal{C}_4$) & $84.1$ & $65.2$ & $2.9$M
\\
RREDet-s ($\mathcal{C}_4$) & $86.3$ & $67.6$ & $10.5$M
\\
RREDet-m ($\mathcal{C}_4$) & $\textbf{87.4}$ & $\textbf{70.3}$ & $25.7$M
\\
\bottomrule
\end{tabular}
\caption{Average Precise (AP) on the PASCAL VOC07+12 dataset. All models are trained from scratch.}
\label{tab3}
\end{table}

\begin{table}[htpb]
\centering
\begin{tabular}{lccc}
\toprule
Model & $\text{AP}_{\text{50}}$ & $\text{AP}_{\text{50:95}}$ & \#Param.
\\
\midrule
YOLOv5-s & $\textbf{56.8}$ & $37.4$ & $7.2$M
\\
YOLOv6-n & $53.1$ & $37.5$ & $4.7$M
\\
YOLOv7-tiny & $55.2$ & $37.4$ & $6.2$M
\\
YOLOv8-n & $52.6$ & $37.3$ & $3.2$M
\\
YOLOv9-n & $53.1$ & $38.3$ & $2.0$M
\\
YOLOv10-n & - & $38.5$ & $2.3$M
\\
\midrule
RREDet-n ($\mathcal{C}_4$) & $55.2$ & $\textbf{40.2}$ & $3.1$M
\\
\bottomrule
\end{tabular}
\caption{Average Precise (AP) on the MS COCO2017 dataset. All models are trained from scratch.}
\label{tab4}
\end{table}

\subsubsection{Comparison with other models in 2D Object Detection.} 
As shown in Table \ref{tab3}, three different-size RRENet models (i.e., RREDet-n / m / s) on the group $\mathcal{C}_4$ mainly compares with the advanced YOLOv8 with three size models (i.e., YOLOv8-n / m / s) on the PASCAL VOC07+12 dataset. Among them, RREDet-n ($\mathcal{C}_4$) achieves approximate $\text{AP}_{\text{50}}$ and $\text{AP}_{\text{50:95}}$ compared to YOLOv8-m, but its parameters are only $11\%$ of YOLOv8-m. In addition, RREDet-s / m ($\mathcal{C}_4$) exceeds other models in both $\text{AP}_{\text{50}}$ and $\text{AP}_{\text{50:95}}$. Although RREDet-m ($\mathcal{C}_4$) has slightly higher $\text{AP}$ metric than RREDet-s ($\mathcal{C}_4$), its parameters are $2.45\times$ that of RREDet-m ($\mathcal{C}_4$). RREDet-s ($\mathcal{C}_4$) balances $\text{AP}$ and parameters.

Furthermore, we conduct experiments on the larger-scale MS COCO2017 dataset to test the proposed RREDet models' generalization ability. 
We compare RREDet-n ($\mathcal{C}_4$) with other YOLO family models of the same scale, as shown in Table \ref{tab4}. From the table, RREDet-n ($\mathcal{C}_4$) outperforms the other models mentioned above in $\text{AP}_{\text{50:95}}$, but is lower than YOLOv5-s in $\text{AP}_{\text{50}}$. Nevertheless, the parameters of YOLOv5-s are $2\times$ that of RREDet-n ($\mathcal{C}_4$). The latest models YOLOv9-n / YOLOv10-n have approximately $65\%$ / $74\%$ parameters of RREDet-n ($\mathcal{C}_4$), but RREDet-n ($\mathcal{C}_4$) improve around $5\%$ / $4.4\%$ in $\text{AP}_{\text{50:95}}$. 

\subsubsection{Summary.} The experiments above prove the advancement of the proposed method for RRE in 2D vision. The RSB is common in the real world. Therefore, relaxing SRE to obtain RRE is an effective way to adapt to RSB. Since the proposed $G$-Biases are learnable parameters that can be updated end-to-end during the training period, they are automatically updated based on the distribution characteristics of the natural dataset to adapt to RSB. In Pytorch, we use $\textit{torch.nn.Parameter}(\cdots, \textit{requires\_grad=True})$ to define $G$-Biases that can be updated by gradient descent direction. Overall, RRE networks can achieve better results than traditional SRE networks in 2D vision.

\subsection{Visualization of SRE, RRE, and NRE.}
\begin{figure}[t]
\centering
\includegraphics[width=1\columnwidth]{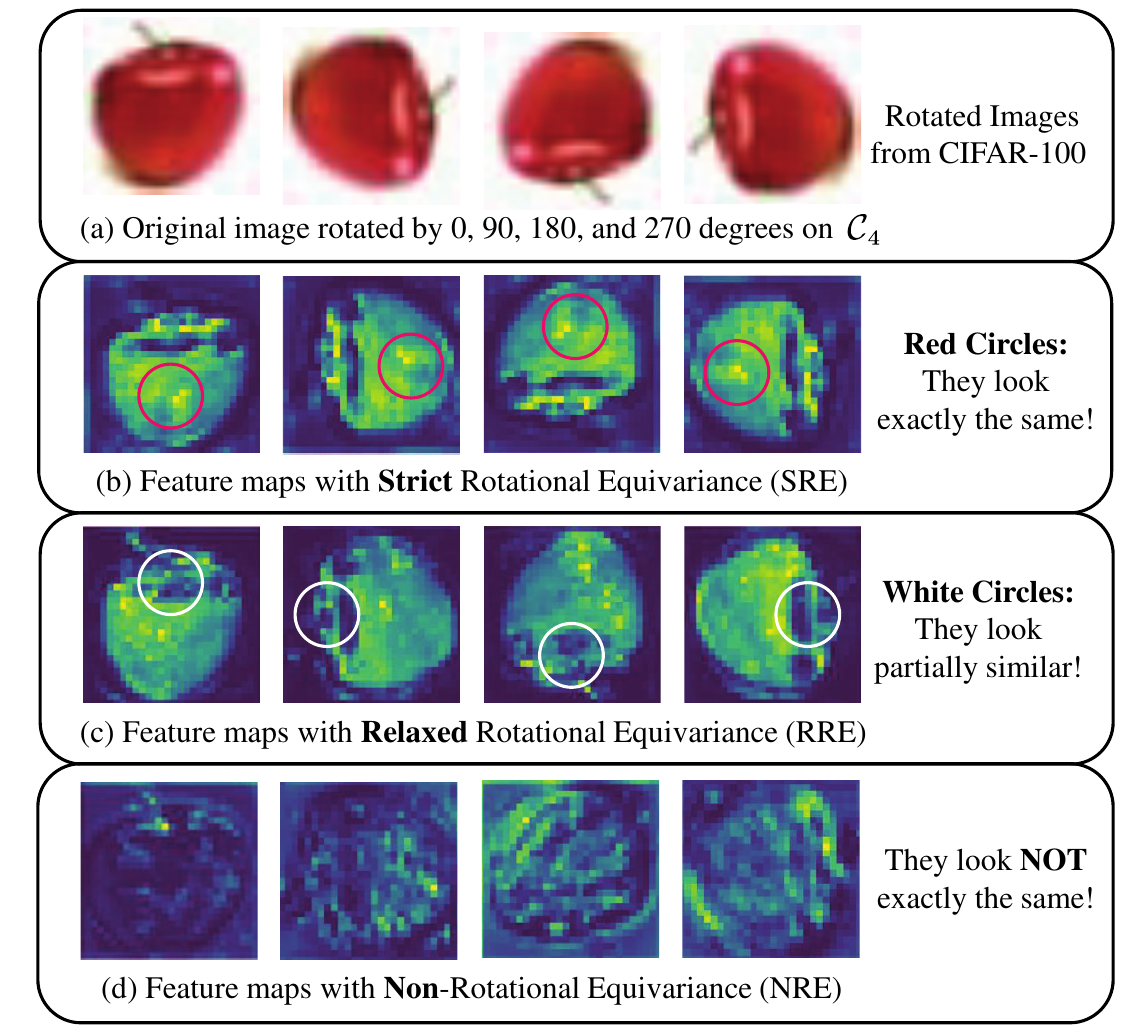}
\caption{The visualization of SRE, RRE, and NRE.}
\label{fig6}
\end{figure}

The visualization of SRE, RRE, and RRE on the group $\mathcal{C}_4$ can be seen in Figure \ref{fig6}.
In (b), we observe that the content inside red circles remained unchanged after rotation, and the overall image shows the same, reflecting SRE's characteristics. In (c), we find slight differences in the content inside white circles after rotation, but the overall image presents similarity, reflecting RRE's characteristics.
In (d), they are almost different in the content. The similarity of RRE, rather than uniformity, is more common and primarily reflects the RSB of objects in the real world.

\section{Conclusion}
This paper delves into rotational equivariant networks for modelling natural datasets, highlighting their effectiveness in leveraging rotation groups. Despite their advancements, existing networks capture uniform and strict rotational symmetry from original data, which does not align with real-world data characterized by {Rotational Symmetry-Breaking} (RSB). This inability to effectively adapt RSB scenarios within natural datasets necessitates a novel method. 

To tackle this challenge, we introduce a simple yet powerful method involving a set of learnable parameters called $G$-Biases. Utilizing this innovative mechanism, we propose a {Relaxed Rotational Equivariant Convolution} (RREConv), tailored to address the nuances of RSB. The extensive experiments have proven the effectiveness of our method.

Exploring Symmetry-Breaking with relaxed equivariance within the realms of 2D and 3D visual fields represents a compelling avenue for future research. We firmly believe incorporating this relaxation principle into strict equivariant models can improve their ability to represent Symmetry-Breaking phenomena in real-world contexts.

\section{Acknowledgments}
This research is supported by the National Natural Science Foundation of China (No.42130112, No.42371479), General Program of Shanghai Natural Science Foundation(Grant No.24ZR1419800, No.23ZR1419300), Science and Technology Commission of Shanghai Municipality (Grant No.22DZ2229004), Beijing Natural Science Foundation (No.QY23187), and Shanghai Frontiers Science Center of Molecule Intelligent Syntheses.
\bibliography{aaai25}

\end{document}